\documentclass[letterpaper, 10 pt, conference]{ieeeconf}  

\IEEEoverridecommandlockouts                              

\usepackage{cite}
\usepackage{amsmath,amssymb,amsfonts}
\usepackage{algorithmic}
\usepackage{graphicx}
\usepackage{textcomp}
\usepackage{multirow} 
\usepackage{xcolor}
\usepackage{comment}

\title{\LARGE \bf
SAMKD: Spatial-aware Adaptive Masking Knowledge Distillation for Object Detection
}

\author{Zhourui Zhang, Jun Li$^*$, Jiayan Li and Jianhua Xu
\thanks{$^{*}$Corresponding author (lijuncst@njnu.edu.cn). All authors are with the School of Computer and Electronic Information,  Nanjing Normal University, Nanjing 210023, China. This work was supported by the National Natural Science Foundation of China under Grant 62173186, 62076134, 62303230.}
}

\begin{document}

\maketitle
\thispagestyle{empty}
\pagestyle{empty}

\begin{abstract}

Most of recent attention-guided feature masking distillation methods perform knowledge transfer via global teacher attention maps without delving into fine-grained clues. Instead, performing distillation at finer granularity is conducive to uncovering local details supplementary to global knowledge transfer and reconstructing comprehensive student features. In this study, we propose a Spatial-aware Adaptive Masking Knowledge Distillation (SAMKD) framework for accurate object detection. Different from previous feature distillation methods which mainly perform single-scale feature masking, we develop spatially hierarchical feature masking distillation scheme, such that the object-aware locality is encoded during coarse-to-fine distillation process for improved feature reconstruction. In addition, our spatial-aware feature distillation strategy is combined with a masking logit distillation scheme in which region-specific feature difference between teacher and student networks is utilized to adaptively guide the distillation process. Thus, it can help the student model to better learn from the teacher counterpart with improved knowledge transfer and reduced gap. Extensive experiments for detection task demonstrate the superiority of our method. For example, when FCOS is used as teacher detector with ResNet101 backbone, our method improves the student network from 35.3\% to 38.8\% mAP, outperforming state-of-the-art distillation methods including MGD, FreeKD and DMKD.

\end{abstract}

\section{INTRODUCTION}

\begin{figure}[ht]
\centering
\includegraphics[width=0.48\textwidth]{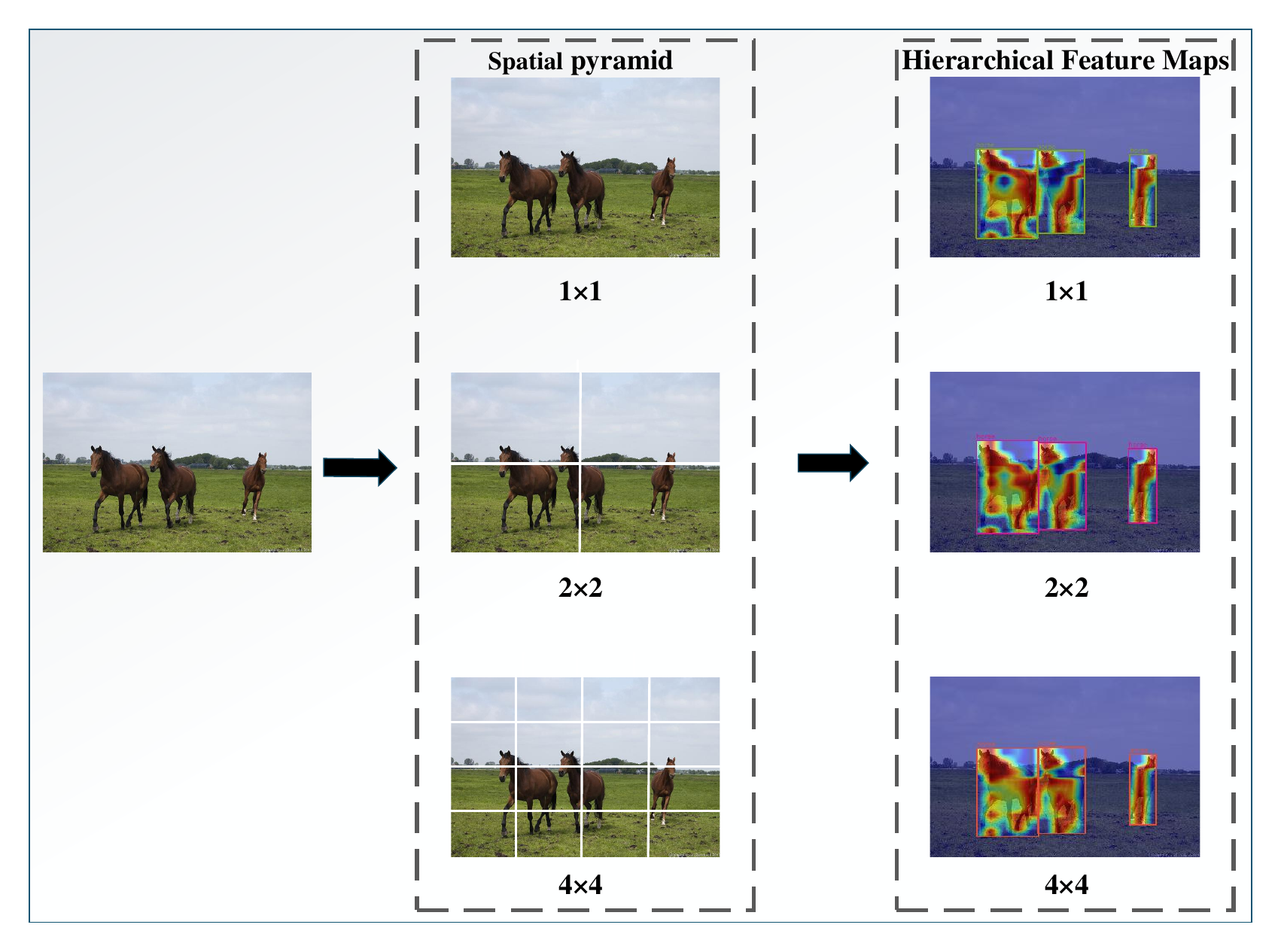}
\caption{Illustration of feature maps obtained by our proposed SAMKD. Benefiting from the spatially hierarchical distillation mechanism, our model can encode region-specific semantics at different feature granularities, significantly improving object awareness and facilitating model adaptability. Specifically, when detecting multiple horses in the image, our model can focus on their bodies and legs at finer $4\times4$ scale, leading to the feature heatmap with high responses to the central regions.}
\label{fig1}
\end{figure}

Knowledge distillation (KD) is capable of improving the performance of a small model (student) with the help of a larger network (teacher) via effective knowledge transfer. It allows the deployment of the student model in resource-constrained scenarios with lower computational budgets~\cite{r1,r2,r4,r5,r26}. Known as logit distillation (LD), numerous LD methods function by aligning the logits between dual networks, such that the student can directly learn from the teacher model for accurate classification~\cite{r16,r17}. In addition, another major line of research in KD methods is feature distillation (FD) which helps the student to generate discriminating representations by imitating the teacher feature maps~\cite{r7,r20}. Different from conventional FD methods, recent feature masking distillation schemes replace feature imitation with feature reconstruction. Taking advantage of masking mechanism, discriminative feature encoding can be reconstructed from the selectively masking regions of the student feature maps, which significantly boosts the feature learning capability of the student model~\cite{r20}. In particular, attention-guided masking strategies are helpful for exploring abundant semantic-aware clues, and thus tremendously promote FD performance in a variety of downstream tasks including object classification and detection~\cite{r6,r7,r29,r30}.

In object detection, imbalance between foreground and background regions poses great challenges to detection accuracy, since the background pixels in an image usually outnumber the foreground counterparts~\cite{r3}. Thus, some important clues are likely to be downplayed especially when the student network typically mimics the features built on all the image pixels without adequately focusing on the object-related regions. In recent years, massive efforts are devoted to addressing the issue for improving the detection performance. FGD~\cite{r5} utilizes the ground-truth bounding boxes to separate the foreground and background. SKD~\cite{r4} makes use of attention mechanisms to assign attention weights to each pixel, yielding higher attention values corresponding to foreground objects. Similar approaches including AMD~\cite{r6} and DMKD~\cite{r7} perform attention-guide feature masking to reconstruct the most useful features. Although these methods can effectively propagate the majority of teacher knowledge, they mainly perform global feature masking distillation without taking into account spatial-aware locality during the distillation process. Due to significant variances in object scales and positions, global feature distillation inevitably introduces noises from the background regions and fails to capture object-aware local clues, which adversely affects comprehensive knowledge transfer from the teacher to the student. To mitigate this limitation, we propose a Spatial-aware adaptive masking knowledge distillation (SAMKD) framework for object detection. In particular, we devise a spatially hierarchical feature masking distillation strategy for improving the spatial perception capability of the student model. In contrast to previous methods in which masking distillation only globally transfers important contents from teacher feature maps to student counterparts, we introduce spatial pyramid mechanism into dual feature spaces and perform coarse-to-fine attention-guided feature masking distillation across different spatial scales. Thus, it helps the student model to focus on global informative regions, while maximally maintaining the spatial-aware locality. In addition, region-specific logit distillation is also achieved with the help of adaptive masking based on the spatial feature difference, such that the disparity between the teacher and the student models can be further reduced. Fig.~\ref{fig1} illustrates feature maps generated from our SAMKD. It is shown that our method can accurately characterize global contents maintaining substantial local details by performing spatial-aware distillation. To summarize, the contributions of our work are threefold as follows:
\begin{itemize}
    \item Different from existing masking distillation methods which mostly transfer global feature maps, we introduce the spatial pyramid mechanism and perform coarse-to-fine distillation to encode object-aware locality. With the help of this spatially hierarchical distillation mechanism, our method can help the student model to comprehensively identify semantically attentive regions.
    \item To further improve distillation performance, we impose the region-specific adaptive masking on the logit distillation such that the feature difference between the teacher and the student network at different spatial scales are used as auxiliary masking information for benefiting knowledge transfer.
    \item Extensive experiments demonstrate the promise of our method in detection task, suggesting that our method surpasses other state-of-the-art distillation approaches including MGD, FreeKD and DMKD.
\end{itemize}


The remainder of the paper is structured as follows. After reviewing the related work in Section~\ref{sec2} , we will elaborate on our proposed SAMKD in Section~\ref{sec3} and conduct extensive experimental evaluations in Section~\ref{sec4}. The final paper is concluded in Section~\ref{sec5}.

\section{Related Work}\label{sec2}

\subsection{Knowledge Distillation}

Without loss of generality, knowledge distillation falls into two lines of research: feature-based and logit-based distillation approaches. The former mainly focus on feature imitation which allows the student model to learn from its teacher counterpart across intermediate feature layers. For example, Yang et al.~\cite{r5} proposed FGD which allows the student model to learn from critical local regions and global knowledge of the teacher network through focal and global distillation respectively, thereby achieving accurate object detection by effectively separating object-aware foreground from background regions. Cao et al.~\cite{r8} proposed PKD distillation which leverages Pearson correlation coefficients for feature mimicking to focus on relational information from the teacher while relaxing the constraints on feature magnitude. Recent research suggests that feature reconstruction driven by masking mechanism is preferable to feature imitation, which significantly contributes to improving learning capability of the student model. In particular, attention-based feature masking helps to identify semantically important regions, leading to reconstructed feature encoding with improved discriminating capability~\cite{r6,r7}.

Different from feature-based distillation approaches, LD allows the student to learn from the teacher at the logit level, thereby facilitating logit mimicking for class-specific knowledge transfer. Successfully applied to classification task, the earliest LD method was proposed to reduce the prediction difference~\cite{r17}. Zheng et al.~\cite{r9} transferred knowledge from the classification head to the localization head in object detection and proposed a novel distillation mechanism called localization distillation. The milestone work demonstrated that logit mimicking beats feature imitation, revealing that distilling knowledge of object categories and locations should be handled separately. Another representative method is DKD~\cite{r10} which decouples the classical KD loss into two parts, i.e., target class loss and non-target class loss. By handling them separately, the flexibility of balancing the two parts is significantly improved and state-of-the-art LD performance is achieved for accurate classification.

\begin{figure*}[ht]
\centering
\includegraphics[width=0.96\linewidth]{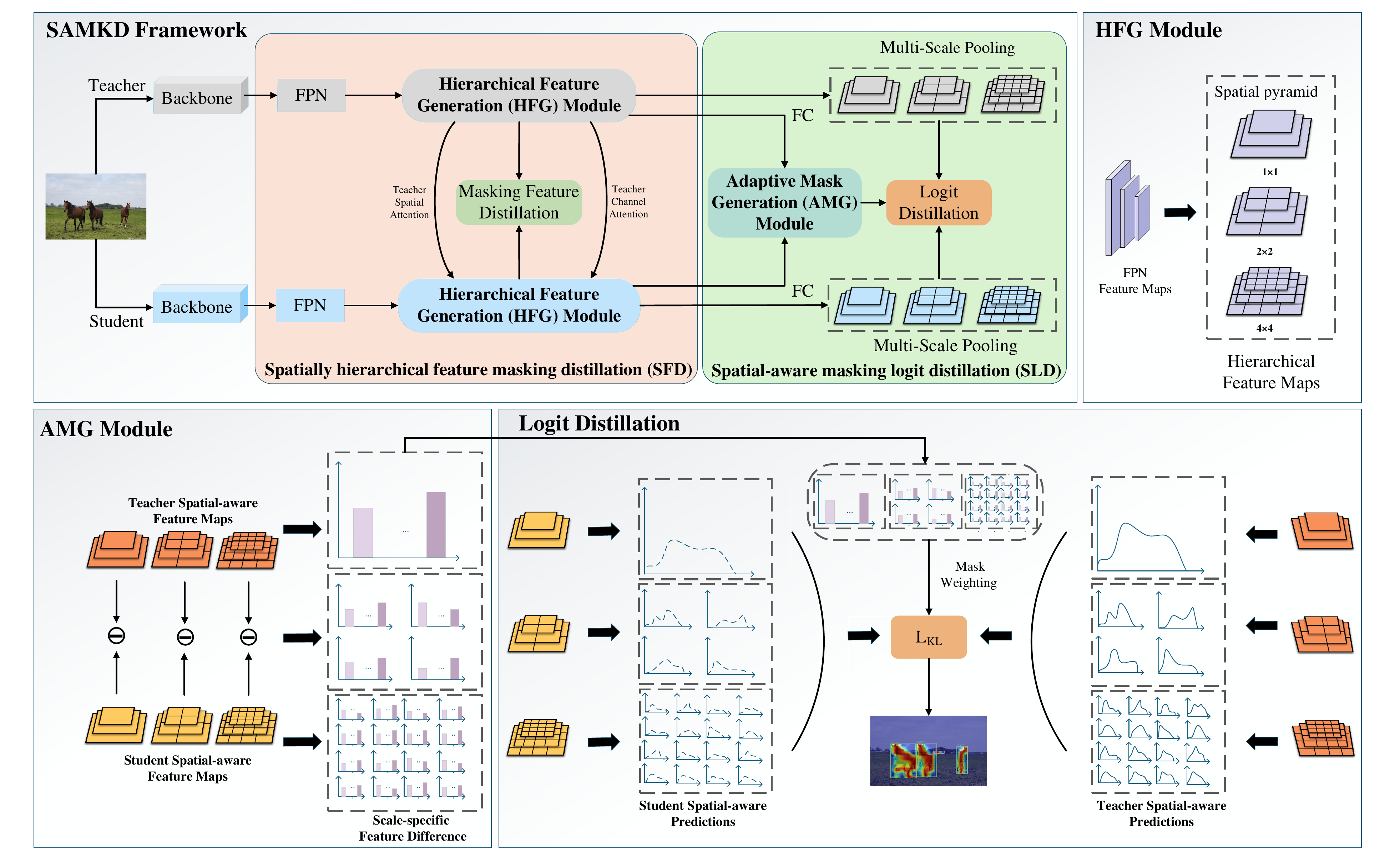}
\caption{Illustration of our proposed SAMKD framework. For both the teacher and the student models, spatial pyramid mechanism is integrated into feature pyramid networks (FPN) from scratch and a spatial-aware feature masking distillation scheme is developed. Specifically, the masking regions on the student feature maps are obtained by dual attention maps generated from the teacher model at each level and comprehensive student feature reconstruction can be achieved via hierarchical feature distillation. Furthermore, the differences between the teacher and the student feature maps within respective divided regions are adopted as auxiliary masking weights for facilitating coarse-to-fine logit distillation, leading to accurate predictions of the student logits.}
\label{fig2}
\end{figure*}

\subsection{Object Detection}

As a mainstream vision task, object detection has received longstanding attention and related research has prospered over decades. Earliest detection models are mainly anchor-based two-stage detectors including the family of R-CNN networks~\cite{r11,r25} which operate by detecting potential objects based on region proposals. Emerging as more efficient detectors, one-stage architectures are considered to be desirable alternative to two-stage methods, since the former is essentially an ensemble model which integrates region proposal generation and detection into a unified framework. In particular, representative YOLO-series models have dramatically advanced the research on object detection, thereby achieving unprecedented success~\cite{r12,r31}. Different from anchor-based models, anchor-free models enjoy better flexibility, since they are point-based detectors independent of the predefined anchor boxes~\cite{r13,r32}. In recent years, a more powerful detector based on Vision Transformer (ViT) has become another popular line of research~\cite{r14}. It allows end-to-end detection based on optimally regressed bounding box instead of involving the post-processing step, i.e., Non maximum Suppression (NMS), and achieves higher detection accuracy. Benefiting from Natural Language Processing (NLP) similar to ViT, detectors based on emerging Vision Mamba architectures introduce bi-directional state space model to encode contextual clues~\cite{r15,r33,r34}. Despite effective, the above-mentioned detectors have suffered complex network structures and expensive training costs, making it difficult to deploy them in resource-constrained scenarios. Consequently, it is indispensable for distilling the knowledge from the heavyweight models to lightweight detectors, such that the latter facilitates real-world applications with improved performance. 

\section{Methods}\label{sec3}

While the existing distillation model guided by attention-based feature masking mechanism allows the student to focus on discriminative regions by globally learning from the teacher model, it suffers limited knowledge transfer without sufficiently exploring fine-grained local details. Considering significant variances in object scale and distributions, encoding spatial information into feature masking distillation can dramatically contribute to masking feature reconstruction with enhanced object awareness. Towards this end, we propose a spatial-aware adaptive masking knowledge distillation method termed SAMKD as illustrated in Fig.~\ref{fig2}. Different from the existing approaches that achieve attention-guided feature reconstruction via global distillation~\cite{r6,r7}, we introduce the spatial pyramid mechanism into the feature masking distillation framework, allowing region-specific feature alignments between the student and the teacher networks. More specifically, we generate coarse-to-fine hierarchical feature maps from feature pyramid networks (FPNs), and thereafter perform spatial-aware feature masking and reconstruction hierarchically. Thus, our scheme simultaneously encodes important global clues and local details, achieving comprehensive distillation. Furthermore, we utilize the region-specific feature difference between the student and the teacher as auxiliary masking weights for aligning the student and teacher logits, such that biased predictions produced from the student is prioritized in the coarse-to-fine logit distillation. Next, we will elaborate on the above-mentioned components in our SAMKD framework. 


\subsection{Spatially hierarchical feature masking distillation}\label{AA}

\begin{figure*}[ht]
\centering
\includegraphics[width=0.96\linewidth]{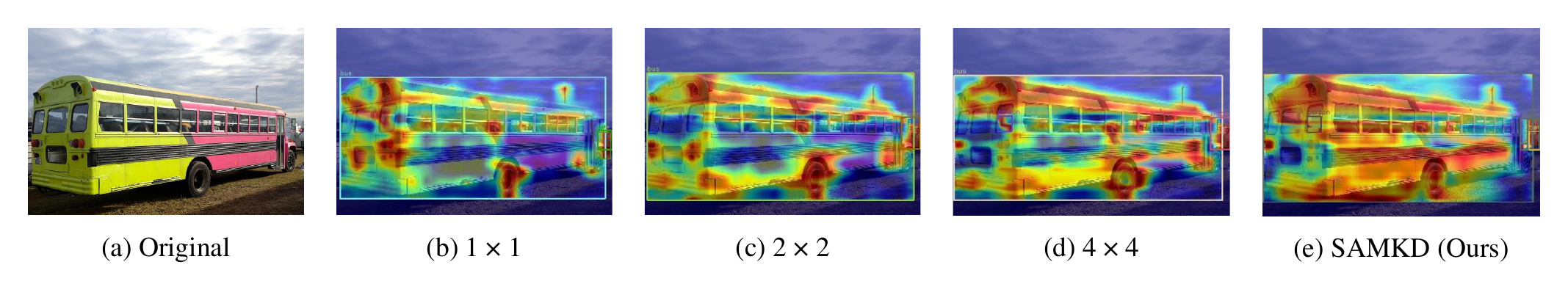}
\caption{Visualizations of different feature maps. Given the original image (a), our coarse-to-fine distillation method (e) can capture more critical object-aware clues (e.g., bus roof and side) compared with performing distillation on individual scale $1\times1$ (b), $2\times2$ (c), and $4\times4$ (d), and thus achieves more accurate detection results. In implementation, the teacher and the student models are GFL-ResNeXt101 and GFL-ResNet50.}
\label{fig3}
\end{figure*}

In the first place, we introduce spatial pyramid mechanism into FPNs of both the teacher and the student, leading to hierarchical feature maps of multiple regions with varying sizes. The spatial pyramid in our framework has three coarse-to-fine hierarchical levels with respective $1\times1$, $2\times2$, and $4\times4$ divided regions. Due to significant variances in object sizes and distributions, this spatial decomposition is conducive to maintaining object-aware locality, allowing the student model to focus on both global contents and finer details for collaborative distillation. 

Next, we leverage attention-guided masking strategy for propagating teacher knowledge and reconstructing student features across different spatial scales. The resulting spatial-aware features will be aggregated for feature distillation, while teacher-student feature difference will be used as auxiliary weighting information for subsequent adaptive logit distillation. Consistent with~\cite{r7}, we leverage dual attention mechanism for identifying attentive regions, such that comprehensive feature reconstruction and distillation can be achieved. To begin with, both spatial and channel attention maps are derived from the teacher features as follows:

\begin{equation}\label{eq:6}\footnotesize
A^s(x,y) =\phi (Sigmoid(\frac{1}{\tau C} \left \langle   \left \| F_{1}^{T} (x,y)\right \|_{2}^{2} , \cdot \cdot \cdot , \left \| F_{n}^{T} (x,y)\right \|_{2}^{2}\right \rangle) 
\end{equation}

\begin{equation}\label{eq:7}\scriptsize
A^c(x,y) =Sigmoid(\frac{1}{\tau H_RW_R}\sum_{h=1}^{H_R}\sum_{w=1}^{W_R}\left \langle   F^T_{h,w,1}(x,y) ,\cdot \cdot \cdot ,F^T_{h,w,C}(x,y)\right \rangle)
\end{equation}
where $A^s(x,y)\in\mathbb{R}^{H_R \times W_R \times 1}$ and $A^c(x,y)\in\mathbb{R}^{1 \times 1 \times C}$ are spatial and channel attention maps obtained at $R(x,y)$ which is $x_{th}$ cell of $y_{th}$ level. $H_R$ and $W_R$ are respectively height and width of $R(x,y)$, while $C$ is the number of channels. In addition, ${F_i(x,y)} \in \mathbb{R}^{C\times1\times1} (i=1,\cdots,N)$ is pixel-wise feature embedding within $R(x,y)$, while $F_{h,w,c}\in\mathbb{R}^{1\times1\times1}$ represents the feature at coordinates $(w, h)$ of $c_{th}$ channel. The hyper-parameter $\tau$ is introduced to adjust the distribution~\cite{r17}. With the produced dual attention maps $A^{s}$ and $A^{c}$, the corresponding masking maps $M^{s}$ and $M^{c}$ can be obtained by:

\begin{equation}
M^s_{i,j}(x,y)=\begin{cases}
  0& A^s_{i,j}(x,y) \ge \omega^s \\
  1& Otherwise
\end{cases}
\end{equation}

\begin{equation}
M^c_k(x,y)=\begin{cases}
  0& A^c_k(x,y) \ge \omega^c \\
  1& Otherwise
\end{cases}
\end{equation}
where $\omega^s$ and $\omega^c$ are predefined masking thresholds. Then, we impose spatial and channel masking on the original student feature map $F_S(x,y)$ with $M^{s}$ and $M^{c}$ formulated as:

\begin{equation}
F_{s}^{S} (x,y)=\Theta (F_{S} (x,y) )\odot M^{s} (x,y)
\end{equation}

\begin{equation}
F_{c}^{S} (x,y)=\Theta (F_{S} (x,y) )\odot M^{c} (x,y)
\end{equation}
where $\Theta(\cdot)$ represents the feature mapping, while $\odot$ denotes Hadamard product. For student feature reconstruction, we respectively utilize convolution-based and perceptron-based blocks to rebuild the masked spatial and channel-aware features~\cite{r7}. The resulting student features are obtained by combining dual reconstructed features as:

\begin{equation}\label{eq:recon}
F_{recon}^{S} (x,y)=\lambda \cdot \theta _{s} (F_{s}^{S}(x,y) )+\mu  \cdot \theta _{c} (F_{c}^{S}(x,y) )
\end{equation}
where $\lambda$ and $\mu$ are two balancing weights, while $\theta_{s}(\cdot)$ and $\theta_{c}(\cdot)$ represent the convolution and perceptron reconstruction blocks, respectively. Thus, final feature distillation is performed between the recovered student features $F_{recon}^{S}$ and the corresponding teacher features $F^T$, and hierarchical knowledge transfer is achieved by aggregating varying-scale distillation losses:

\begin{equation}
    L_{Feat} =\sum_{y}\sum_{x} \|F^T(x,y)- \Theta(F_{recon}^S(x,y) \|_2^2
\end{equation}
where $F^T(x,y)$ denote the original feature maps generated from the teacher network within $R(x,y)$.

Fig.~\ref{fig3} intuitively compares various feature maps obtained by both single-scale distillation approaches and our coarse-to-fine distillation strategy. Benefiting from integrating reconstructed features at multiple spatial scales, our method can capture both coarse-grained visual cues and fine-grained local details, thereby generating more distinguishable features with improved object awareness.

\subsection{Spatial-aware adaptive logit distillation}

In addition to the above-mentioned spatially hierarchical feature masking distillation, spatial pyramid modeling is also introduced into logit distillation such that spatial-aware knowledge transfer at the logit level can be achieved. In particular, the difference between the teacher and the student feature maps across varying spatial scales is calculated and utilized as auxiliary masking weights for guiding logit distillation. The resulting spatial masks not only maintain object spatial awareness but also facilitates adaptive logits transfer of the teacher, which further narrows the logits gap between the teacher and the student. More specifically, scale-specific average pooling is applied to the logits for better aligning the prediction results of the teacher and the student. Meanwhile, it somewhat suppresses the effects of background prediction in logit distillation, which benefits the object-specific knowledge transfer.

Mathematically, logit outputs $Z(x,y)\in \mathbb{R}^{1\times1\times K}$ of the region $R(x,y)$ can be calculated by imposing spatial-aware average pooling~\cite{r16} on logits as follows:

\begin{equation}
Z_{T} (x,y)=\sum_{(p,q)\in R(x,y)} \frac{L_{T}(p,q) }{X^{2} } 
\end{equation}

\begin{equation}
Z_{S} (x,y)=\sum_{(p,q)\in R(x,y)} \frac{L_{S}(p,q) }{X^{2} } 
\end{equation}
where $L(p,q)$ denotes the logit score at coordinates $(p,q)$ of $R(x,y)$, while $X$ is the total number of spatial cells at $x_{th}$ level. To further bridge the gap between the teacher and the student, we utilize the difference between the region-specific teacher and student features across varying scales as auxiliary masking weights to facilitate adaptive knowledge transfer. Mathematically, it can be formulated as 

\begin{equation}
W(x,y)=diff(F_{T}(x,y),F_{S}(x,y))
\end{equation}
where $diff(u,v)$ can be formulated as cosine distance metric $1 - \frac{u^{T}v}{\|u\|_2 \|v\|_2} \in [0, 2]$. Thus, logit distillation loss can be built via weighted summation of the final teacher and student score maps $Z_{S/T}(x,y)$ across varying spatial scales with $W(x,y)$:

\begin{equation}
L_{Logit} =\sum_{x,y} W(x,y)\cdot D(Z_{T} (x,y),Z_{S} (x,y))
\end{equation}
where $D(\cdot)$ is the logit distillation loss function formulated as $KL$ loss in our method.

\subsection{Loss Function}

For training our SAMKD, overall loss function can be formulated as:

\begin{equation}\label{eq:16}
L_{SAMKD} = L_{Det} + \alpha L_{Feat} + \beta L_{Logit}
\end{equation}
where $L_{Det}$ is the original detection loss, while $L_{Feat}$ and $L_{logit}$ represent the above-mentioned spatial-aware feature and logit distillation losses, respectively. $\alpha$ and $\beta$ are two tradeoff hyperparameters.

\begin{table*}[htbp] 
\caption{Comparison of our method with other distillation methods in COCO dataset (\%). AP/AR$_{s/m/l}$ refers to the detection accuracy with respect to small/medium/large-size objects. The best results are highlighted in bold.}
\normalsize
\begin{center}
\begin{tabular}{c|c|cccc|cccc}
\hline
Teacher                                                                      & Student        & mAP                 & AP$_s$           & AP$_m$           & AP$_l$           & mAR           & AR$_s$           & AR$_m$           & AR$_l$           \\ \hline
\multirow{6}{*}{\begin{tabular}[c]{@{}c@{}}GFL   \\ ResNet101\end{tabular}}  & GFL-ResNet50            & 39.6                & 22.3          & 43.6          & 52.2          & 58.2          & 35.8          & 63.1          & 75.2          \\
                                                                             & +FGD            & 40.1(+0.5)          & 23.1          & 44.0          & 53.1          & 58.7          & 37.4          & 63.4          & 75.5          \\
                                                                             & +MGD            & 40.3(+0.7)          & 22.9          & 44.1          & 53.6          & 59.1          & 37.2          & 63.6          & 76.2          \\
                                                                             & +FreeKD         & 41.4(+1.8)          & 23.5          & 45.7          & 54.2          & 59.7          & 38.4          & 64.6          & 75.4          \\
                                                                             & +DMKD           & 41.8(+2.2)          & 23.6          & 46.2          & 54.6          & 60.1          & 38.6          & 64.8          & \textbf{76.1} \\
                                                                             & \textbf{+SAMKD(Ours)} & \textbf{42.1(+2.5)} & \textbf{24.2} & \textbf{46.5} & \textbf{54.7} & \textbf{60.2} & \textbf{38.9} & \textbf{64.8} & 75.8          \\ \hline            
\multirow{6}{*}{\begin{tabular}[c]{@{}c@{}}ATSS \\ ResNet101\end{tabular}}      & ATSS-ResNet50           & 39.4                & 22.1          & 43.3          & 52.1          & 58.1          & 35.6          & 63.1          & 75.0          \\
                                                                                & +FGD            & 41.0(+1.6)          & 24.1          & 45.5          & 53.4          & 60.1          & 40.1          & 61.5          & 72.8          \\
                                                                                & +MGD            & 41.2(+1.8)          & 24.3          & 45.8          & 53.6          & 60.2          & 40.2          & 61.7          & 73.0          \\
                                                                                & +FreeKD         & 41.3(+1.9)          & 24.3          & 45.9          & 53.7          & 60.2          & 40.4          & 61.7          & 72.9          \\
                                                                                & +DMKD           & 41.6(+2.2)          & 24.6          & 46.1          & 53.9          & 60.4          & 40.5          & 65.6          & 76.7          \\
                                                                            & \textbf{+SAMKD(Ours)} & \textbf{41.8(+2.4)} & \textbf{24.8} & \textbf{46.2} & \textbf{53.9} & \textbf{60.5} & \textbf{40.6} & \textbf{65.7} & \textbf{77.1} \\ \hline
\multirow{6}{*}{\begin{tabular}[c]{@{}c@{}}RetinaNet\\ ResNet101\end{tabular}}  & RetinaNet-ResNet50      & 37.4                & 20.6          & 40.7          & 49.7          & 53.9          & 33.1          & 57.7          & 70.2          \\
                                                                                & +FGD            & 38.7(+1.3)          & 20.5          & 43.1          & 52.3          & 55.5          & 34.7          & 60.2          & 72.1          \\
                                                                                & +MGD            & 38.9(+1.5)          & 20.9          & 43.4          & 52.8          & 55.8          & 35.2          & 60.5          & 72.3          \\
                                                                                & +FreeKD         & 39.0(+1.6)          & 20.6          & 43.2          & 52.7          & 55.9          & 35.3          & 60.5          & 72.4          \\
                                                                                & +DMKD           & 39.3(+1.9)          & 21.5          & 43.5          & 52.4          & 56.3          & 35.4          & \textbf{60.8} & 72.6          \\
                                                                                & \textbf{+SAMKD(Ours)} & \textbf{39.5(+2.1)} & \textbf{21.7} & \textbf{43.5} & \textbf{52.9} & \textbf{56.4} & \textbf{35.5} & 60.6          & \textbf{72.9} \\ \hline
\multirow{6}{*}{\begin{tabular}[c]{@{}c@{}}FCOS  \\ ResNet101\end{tabular}}  & FCOS-ResNet50            & 35.3                & 20.1          & 38.3          & 46.2          & 53.0          & 32.3          & 57.4          & 69.4          \\
                                                                             & +FGD            & 38.0(+2.7)          & 21.4          & 31.5          & 49.5          & 55.0          & 34.3          & 59.4          & 71.1          \\
                                                                             & +MGD            & 38.2(+2.9)          & 21.5          & 31.8          & 49.7          & 55.3          & 34.6          & 59.7          & 71.5          \\
                                                                             & +FreeKD         & 38.3(+3.0)          & 21.7          & 41.9          & 49.7          & 55.3          & 34.5          & 59.9          & 71.7          \\
                                                                             & +DMKD           & 38.5(+3.2)          & \textbf{22.0} & 42.1          & 49.8          & 55.5          & 34.7          & 60.3          & 71.8          \\
                                                                             & \textbf{+SAMKD(Ours)} & \textbf{38.8(+3.5)} & 21.9          & \textbf{42.3} & \textbf{50.1} & \textbf{55.7} & \textbf{34.9} & \textbf{60.3} & \textbf{71.9} \\ \hline
\end{tabular}
\label{tab1}
\end{center}
\end{table*}

\section{Experiments}\label{sec4}

\subsection{Experimental Setup}\label{AA}
For detection task, all the methods are conducted on public COCO dataset~\cite{r18} which comprises over 320k images of 80 different object categories with abundant annotations. Among them, 120k images are used for training while 5k images for testing in all the experiments. For performance measure, both Average Precision (AP) and Average Recall (AR) are adopted as evaluation metrics following~\cite{r19}. All the experiments are conducted on a desktop with an Intel i9-10900K CPU and a 3090 GPU under PyTorch framework. During the training process, SGD optimizer is used for training all the detectors within 24 epochs. Meanwhile, momentum is set as 0.9 whilst weight decay is set to 0.0001. To demonstrate the superiority of our SAMKD model, numerous state-of-the-art (SOTA) methods are involved in our comparative studies, including FGD~\cite{r5}, MGD~\cite{r20}, DMKD~\cite{r7} and FreeKD~\cite{r21}. Meanwhile, four popular deep detectors including GFL~\cite{r22}, ATSS~\cite{r23}, RetinaNet~\cite{r3} and FCOS~\cite{r13} are utilized as either teacher or student models.

\subsection{Results}

Table~\ref{tab1} compares different distillation methods with various detection models. With the help of GFL teacher detector using ResNet101 backbone, our SAMKD improves the student used as GFL-ResNet50 by 2.5\% accuracy, reporting the highest 42.1\% mAP score. The performance gains achieved by our method consistently surpass other distillation approaches, suggesting that SAMKD outperforms advanced MGD, FreeKD and DMKD by 1.8\%, 0.7\% and 0.3\% mAP respectively. In particular, our approach reports 24.2\% APs and exceeds the three competitors by 1.3\%, 0.7\% and 0.6\% respectively. This substantially demonstrates the promise of SAMKD in uncovering fine-grained clues and promoting detection of small objects with hierarchical distillation mechanism. When the detection framework is replaced by ATSS, SAMKD achieves significant performance boost over the ATSS-ResNet50 by 2.4\% and beats the other competitors across all metrics. Similar performance improvements can also be observed with RetinaNet detector. More specifically, our method allows the student model to achieve 2.1\% mAP improvements, and consistently beats FreeKD and DMKD by 0.5\% and 0.2\%. Furthermore, we have also carried out additional distillation experiments in which FCOS is used for detection framework. It is shown that the proposed SAMKD still exhibits superior performance and improves the student from 35.3\% to the highest 38.8\% mAP. In addition to detection precision, the performance boosts in AR-related metrics are also reported using the aforementioned detectors. These results indicate that our coarse-to-fine distillation strategy is beneficial for mitigating enormous gap between teacher and student models.

\subsection{Ablation studies}

Since the proposed SAMKD distillation framework consists of two primary components, namely spatially hierarchical feature masking distillation and spatial-aware adaptive logit distillation, we will conduct ablation studies in this section to explore the effects of individual module on the model performance. For notation, the two components are abbreviated as \textbf{SFD} and \textbf{SLD}. In practice, we utilize GFL as the teacher network with a more powerful ResNeXt101 backbone, while employ GFL-ResNet50 for the student model. As shown in Table~\ref{tab2}, removing SFD module leads to a performance decline of 0.3\% mAP, suggesting that SFD allows coarse-to-fine knowledge transfer and thus is indispensable for improving the student performance. On the other hand, single SFD module w/o SLD generates 42.2\% mAP which is only slightly inferior to the complete framework incorporating both components. This also indicates that SLD plays a supplementary role in our method and can further reduce the teacher-student disparity. 



\begin{table}[htbp]
\caption{Ablation studies using GFL detector for both the teacher and the student. Two primary components of our proposed SAMKD model are SFD and SLD which are short for spatially hierarchical feature masking distillation and spatial-aware adaptive logit distillation, respectively.}
\normalsize
\begin{center}
\begin{tabular}{ccc}
\hline
\multicolumn{3}{c}{\multirow{2}{*}{\textit{\begin{tabular}[c]{@{}c@{}}Teacher$\colon$ GFL$-$ResNeXt101\\ Student$\colon$ GFL$-$ResNet50\end{tabular}}}} \\
\multicolumn{3}{c}{}                                                                                                                \\ \hline
SFD                              & \multicolumn{1}{c|}{SLD}                              & mAP(\%)                                       \\ \hline
\checkmark                                & \multicolumn{1}{c|}{}                                & 42.2                                      \\
                                 & \multicolumn{1}{c|}{\checkmark }                               & 42.0                                      \\
\checkmark                                 & \multicolumn{1}{c|}{\checkmark }                               & \textbf{42.3}                             \\ \hline
\end{tabular}
\label{tab2}
\end{center}
\end{table}

To have an insight into SFD, we have conducted different experiments to explore the effect of individual distillation scales within SFD. Since three coarse-to-fine spatial scales are involved in SFD, we compare different settings in which both single-scale and multi-scale distillation are considered. Similar to the above-mentioned experimental settings, GFL-ResNeXt101 and GFL-ResNet50 are used for the teacher and the student models, respectively. As demonstrated in Table~\ref{tab3}, complete model incorporating all the three scales reports the highest mAP score of 42.3\%, which is superior to any single-scale distillation strategy with a performance gain of approximately 0.3\%. In particular, single-scale distillation at global $1\times1$ feature map only achieves 41.9\%, which lags behind the complete model by 0.4\%. This implies that fine-grained distillation outweighs the coarse-grained distillation, allowing the student to focus on more local attentive clues with improved performance.

\begin{table}[htbp]
\caption{Performance of our SAMKD with different settings of distillation scales.}
\normalsize
\begin{center}
\begin{tabular}{cccc}
\hline
\multicolumn{4}{c}{\multirow{2}{*}{\textit{\begin{tabular}[c]{@{}c@{}}Teacher$\colon$GFL$-$ResNeXt101\\ Student$\colon$GFL$-$ResNet50\end{tabular}}}} \\
\multicolumn{4}{c}{}                                                                                                                         \\ \hline
S1($1\times1$)                   & S2($2\times2$)                   & \multicolumn{1}{c|}{S3($4\times4$)}                  & mAP(\%) \\
\hline
\checkmark                           &                             & \multicolumn{1}{c|}{}                           & 41.9                           \\
                            & \checkmark                           & \multicolumn{1}{c|}{}                           & 41.9                           \\
                            &                             & \multicolumn{1}{c|}{\checkmark}                          & 42.0                           \\
\checkmark                           & \checkmark                           & \multicolumn{1}{c|}{}                           & 42.0                           \\
\checkmark                           &                             & \multicolumn{1}{c|}{\checkmark}                          & 42.1                           \\
                            & \checkmark                           & \multicolumn{1}{c|}{\checkmark}                          & 42.2                           \\
\checkmark                           & \checkmark                           & \multicolumn{1}{c|}{\checkmark}                          & \textbf{42.3}                  \\ \hline
\end{tabular}
\label{tab3}
\end{center}
\end{table}

Additionally, we have investigated the benefit of performing spatial-aware logits transfer. As shown in Table~\ref{tab4}, our SAMKD using SLD reports superior 42.0\% mAP, exhibiting a 0.2\% improvement over the model adopting conventional logit distillation w/o SLD. This demonstrates that our SLD can adaptively propagate the teacher logits with the help of spatial-aware mask derived from teacher-student feature discrepancy. As a result, region-specific object awareness is further enhanced by paying more attention on the biased predictions and suppressing the background noise to some extent.

\begin{table}[htbp]
\caption{The impact of adaptive logit distillation on our SAMKD. }
\normalsize
\begin{center}
\begin{tabular}{c|cc}
\hline
LD Strategy & w/o SLD & w/ SLD \\ \hline
mAP(\%)                & 41.8              & \textbf{42.0}           \\ \hline
\end{tabular}
\label{tab4}
\end{center}
\end{table}

\subsection{Parameter Analysis}

In this section, we will discuss varying hyperparameters involved in our framework. In terms of $\lambda$ and $\mu$ which are used for balancing spatial and channel-wise feature masking in Eq.~(\ref{eq:recon}), Table~\ref{tab5} reveals that the best result is achieved with $\lambda=0.45$ and $\mu=0.55$, reporting 42.3\% mAP score when transferring from the teacher GFL-ResNext101 to the student GFL-ResNet50. This is also consistent with~\cite{r7}, indicating that a comparatively balanced distribution of dual feature masking contributes to comprehensive feature reconstruction for promoting student performance. 

\begin{table}[htbp]
\caption{Parameter analysis of dual feature masking.}
\normalsize
\begin{center}
\begin{tabular}{ccc}
\hline
\multicolumn{3}{c}{\textit{\begin{tabular}[c]{@{}c@{}}Teacher:GFL-ResNeXt101\\ Student:GFL-ResNet50\end{tabular}}} \\ \hline
$\lambda$                         & \multicolumn{1}{c|}{$\mu$}                         & mAP(\%)                                \\ \hline
0.35                      & \multicolumn{1}{c|}{0.65}                      & 42.2                               \\
0.45                      & \multicolumn{1}{c|}{0.55}                      & \textbf{42.3}                      \\
0.55                      & \multicolumn{1}{c|}{0.45}                      & 42.2                               \\
0.65                      & \multicolumn{1}{c|}{0.35}                      & 42.1                               \\ \hline
\end{tabular}
\label{tab5}
\end{center}
\end{table}

Further investigation of thresholding dual masking regions as shown in Table~\ref{tab6} demonstrates that the best result is achieved with $\omega^{s}=0.95$ and $\omega^{c}=0.50$. Since hierarchical distillation mechanism is employed in our method, higher spatial masking threshold is helpful for capturing fine-grained spatially attentive clues, and thus dramatically benefits coarse-to-fine knowledge transfer. On the other hand, further increasing $\omega^{s}$ is likely to introduce high-frequency background noise, resulting in a slight performance decline of 0.1\% when $\omega^{s}$ increases from 0.95 to 1.05.

\begin{figure}
\centering
\includegraphics[width=0.48\textwidth]{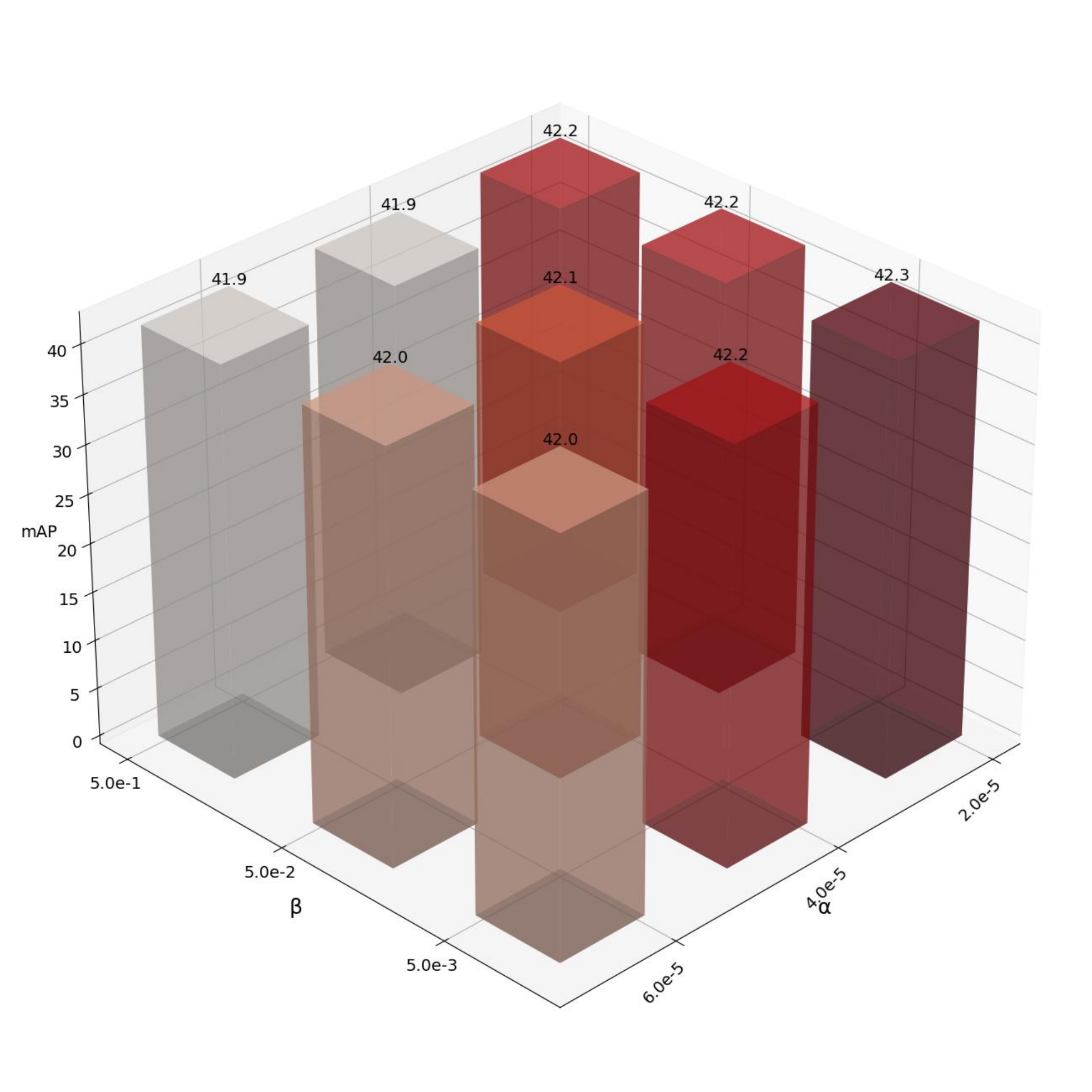}
\caption{Performance of our SAMKD using different values of $\alpha$ and $\beta$ with GFL-ResNeXt101 (teacher) and GFL-ResNet50  (student).}
\label{fig4}
\end{figure}

\begin{table}[htbp]
\caption{Discussion of dual masking thresholds.}
 \normalsize
\begin{center}
\begin{tabular}{cccc}
\hline
\multicolumn{4}{c}{\textit{\begin{tabular}[c]{@{}c@{}}Teacher:GFL-ResNeXt101\\ Student:GFL-ResNet50\end{tabular}}}      \\ \hline
\multicolumn{1}{c|}{$\omega^{s}(\omega^{c}=0.50)$}  & \multicolumn{1}{c|}{mAP(\%)}           & \multicolumn{1}{c|}{$\omega^{c}(\omega^{s}=0.95)$}  & \multicolumn{1}{c}{mAP(\%)} \\ \hline
\multicolumn{1}{c|}{0.95} & \multicolumn{1}{c|}{\textbf{42.3}} & \multicolumn{1}{c|}{0.45} & 42.1                    \\
\multicolumn{1}{c|}{1.00} & \multicolumn{1}{c|}{42.3}          & \multicolumn{1}{c|}{0.50} & \textbf{42.3}           \\
\multicolumn{1}{c|}{1.05} & \multicolumn{1}{c|}{42.2}          & \multicolumn{1}{c|}{0.55} & 42.2                    \\ \hline
\end{tabular}
\label{tab6}
\end{center}
\end{table}

In addition, we also explore collaborative optimization of SFD and SLD in our framework by discussing the weighting coefficients $\alpha$ and $\beta$ in Eq.~(\ref{eq:16}). As shown in Fig.~\ref{fig4}, the best results are obtained when $\alpha$ and $\beta$ are respectively set to $2.0\times10^{-3}$ and $6.0\times10^{-6}$, achieving the best mAP score of 42.3\%.  

\section{Conclusion}\label{sec5}
In this study, we propose a spatial-aware adaptive masking knowledge distillation method termed SAMKD for accurate object detection. Different from the existing masking distillation strategy which globally transfers teacher knowledge, we introduce spatial pyramid mechanism and perform spatially hierarchical feature masking, such that coarse-to-fine feature transfer can be achieved for encoding both global informative contents and fine-grained local attentive clues. In addition, we make use of region-specific feature difference between the teacher and the student as auxiliary masking weights to adaptively perform logit distillation for further bridging the teacher-student gap. Extensive experiments in public COCO dataset demonstrate the promise of our SAMKD which is superior to the state-of-the-art distillation approaches.






\bibliographystyle{elsarticle-num}
\bibliography{root}

\end{document}